\documentclass[10pt,twocolumn,letterpaper]{article}

\usepackage[T1]{fontenc}
\usepackage[utf8]{inputenc}
\usepackage{times}
\usepackage[margin=0.75in]{geometry}
\usepackage{graphicx}
\usepackage{tikz}
\usepackage{amsmath,amssymb}
\usepackage{booktabs}
\usepackage{multirow}
\usepackage{xcolor}
\usepackage{microtype}
\usepackage[colorlinks=true,allcolors=blue]{hyperref}
\usepackage{cleveref}
\usetikzlibrary{arrows.meta,positioning}

\tikzset{
  methodbox/.style={draw,rounded corners=2pt,align=center,inner sep=3pt,font=\scriptsize,minimum height=8mm},
  data/.style={methodbox,fill=yellow!28},
  module/.style={methodbox,fill=blue!8},
  attn/.style={methodbox,fill=orange!25},
  pred/.style={methodbox,fill=green!20},
  arr/.style={-{Stealth[length=1.8mm]},thick}
}

\title{\textbf{S23DR 2026 Winning Solution}}

\author{%
Jan Skvrna \quad Miroslav Purkrabek \quad Lukas Neumann\\
Visual Recognition Group, Czech Technical University in Prague\\
{\tt\small jan.skvrna@cvut.cz \quad purkmir@fel.cvut.cz \quad lukas.neumann@cvut.cz}%
}

\date{}

\begin{document}
\maketitle

\begin{abstract}
This text presents the winning solution to the S23DR 2026 challenge for
structured 3D wireframe reconstruction from sparse SfM, fitted depth, and
semantic segmentations. The method treats vertices as a conditional set and
denoises 64 vertex tokens with a flow-matching DiT conditioned on
Perceiver-style scene tokens. A global pass predicts the coarse structure, a
hull-cropped second pass refines it, and a small multi-sample consensus step
keeps the stochastic sampler well behaved. The final system ranked first on the
private leaderboard, achieving $HSS=0.654$.
\end{abstract}

\section{Introduction}
\label{sec:intro}

The Structured 3D Reconstruction (S23DR) 2026 challenge\footnote{\url{https://huggingface.co/spaces/usm3d/S23DR2026}}~\cite{s23dr2026}
aims to predict the \emph{3D wireframe} of a building given Structure-from-motion reconstruction extended by segmentations for each view. Formally, the target is a labelled 3D graph
$G=(V,E)$ with metric 3D vertices $V=\{v_i\in\mathbb{R}^3\}$ (roof apexes,
eave/ridge endpoints, building corners) and straight edges $E\subseteq V\times V$,
each edge carrying a semantic class defined by its adjacent faces (rake, eave,
flashing, ground line, \ldots). Predictions are scored by the Hybrid Structure Score~\cite{langerman2025hss}, the harmonic mean of a vertex
term and an edge term. The vertex term is the F1 of predicted vs.\ ground-truth
vertices under an optimal (Hungarian) assignment with a metric matching radius;
the edge term is a volumetric IoU in which every edge is rendered as a cylinder
and the predicted and ground-truth edge solids are intersected. Hybrid Structure Score is thus
permutation-invariant and penalizes both missed and spurious structure, rewarding
predictions that recover the correct topology with metrically accurate
vertices and edges.

The task is non-standard because the only inputs are a sparse,
noisy COLMAP point cloud~\cite{schonberger2016sfm}, scale-fitted depth maps, and projected 2D
segmentations~\cite{zhou2017ade20k}, without any RGB images, over highly varied roof geometry
(Sec.~\ref{sec:dataset}). We treat wireframe prediction as conditional set
generation in 3D rather than the discriminative 2D-to-3D lifting used by the
baseline: a fixed set of vertex queries is denoised directly in metric space by a
flow-matching transformer conditioned on a learned scene encoding. The
predictor is run in two stages. First, a global coarse pass and a hull-cropped
refinement pass. To reduce the variance of the prediction multi-sample consensus ensemble selects and refines the
final graph at inference. This approach ranked \textbf{first} on the S23DR~2026
private test leaderboard with an HSS of \textbf{0.654}, ahead of the second-place
entry ($0.648$) and well above the learned ($0.474$) and handcrafted ($0.391$)
organiser baselines; it also achieved the highest vertex F1 ($0.791$) of all
submissions.

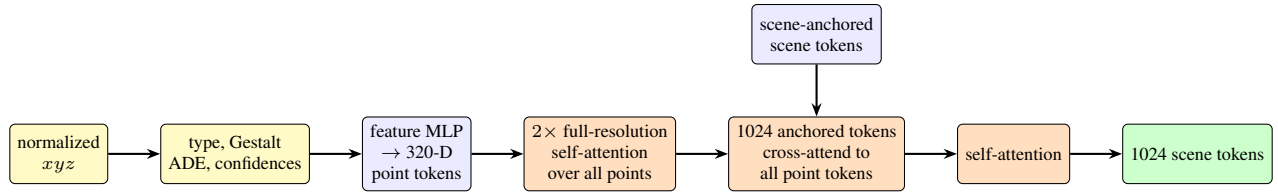
\begin{figure*}[t]
\centering
\resizebox{0.94\textwidth}{!}{%
\begin{tikzpicture}[node distance=7mm and 7mm]
\node[data] (xyz) {normalized\\$xyz$};
\node[data,right=of xyz] (sem) {type, Gestalt\\ADE, confidences};
\node[module,right=of sem] (lift) {feature MLP\\$\rightarrow$ 320-D\\point tokens};
\node[attn,right=of lift] (full) {$2\times$ full-resolution\\self-attention\\over all points};
\node[attn,right=of full] (pool) {1024 anchored tokens\\cross-attend to\\all point tokens};
\node[module,above=of pool] (anchors) {scene-anchored\\scene tokens};
\node[attn,right=of pool] (refine) {self-attention};
\node[pred,right=of refine] (out) {1024 scene tokens};
\draw[arr] (xyz) -- (sem);
\draw[arr] (sem) -- (lift);
\draw[arr] (lift) -- (full);
\draw[arr] (full) -- (pool);
\draw[arr] (anchors) -- (pool);
\draw[arr] (pool) -- (refine);
\draw[arr] (refine) -- (out);
\end{tikzpicture}}
\caption{\textbf{Scene encoder.} Heterogeneous per-point features are lifted to
320-D tokens, contextualized at full resolution, and pooled into a fixed bank of
1024 scene tokens. Pooling tokens are anchored at real scene points selected by
priority-aware Gestalt FPS before they cross-attend to all point tokens.}
\label{fig:encoder}
\end{figure*}

\begin{figure*}[t]
\centering
\resizebox{0.65\textwidth}{!}{%
\begin{tikzpicture}[node distance=5mm and 5mm, every node/.append style={font=\footnotesize}]
\node[data] (slots) {vertex tokens\\$x_t \in \mathbb{R}^{64\times 3}$};
\node[module,right=of slots] (slotemb) {$xyz$ Fourier\\token projection};
\node[data,above=of slotemb] (time) {flow time\\$t\in[0,1]$};
\node[module,right=of time] (temb) {sinusoidal time\\embedding};
\node[data,below=of slotemb] (scene) {scene tokens\\from encoder};
\node[attn,right=of slotemb] (block) {$N\times$ DiT Blocks\\cross-attend to\\scene tokens};
\node[pred,right=of block] (heads) {velocity, validity\\edge logits};
\node[pred,right=of heads] (step) {Euler update};
\draw[arr] (slots) -- (slotemb);
\draw[arr] (slotemb) -- (block);
\draw[arr] (time) -- (temb);
\draw[arr] (temb) -- (block);
\draw[arr] (scene) -- (block);
\draw[arr] (block) -- (heads);
\draw[arr] (heads) -- (step);
\end{tikzpicture}}
\caption{\textbf{Vertex denoiser.} The denoiser updates 64 vertex tokens along
flow time $t\in[0,1]$. DiT blocks~\cite{peebles2023dit} condition the tokens on the time embedding
and encoded scene tokens; the heads output velocity, validity, and pairwise edge
logits used by the Euler update.}
\label{fig:denoiser}
\end{figure*}
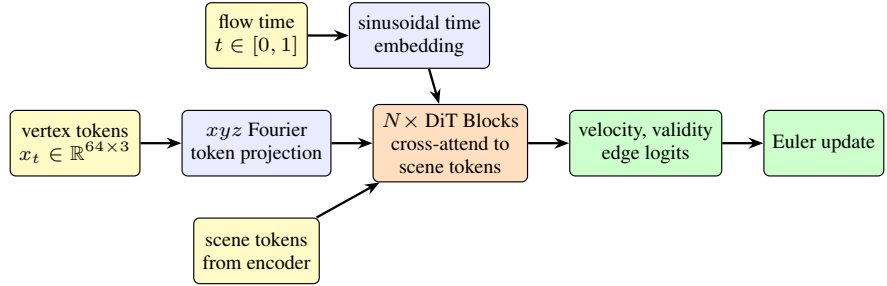

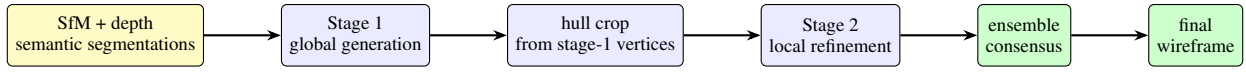
\begin{figure*}[t]
\centering
\resizebox{0.92\textwidth}{!}{%
\begin{tikzpicture}[node distance=10mm and 10mm]
\node[data] (raw) {SfM + depth\\semantic segmentations};
\node[module,right=of raw] (s1) {Stage 1\\global generation};
\node[module,right=of s1] (crop) {hull crop\\from stage-1 vertices};
\node[module,right=of crop] (s2) {Stage 2\\local refinement};
\node[pred,right=of s2] (ens) {ensemble\\consensus};
\node[pred,right=of ens] (out) {final\\wireframe};
\draw[arr] (raw) -- (s1);
\draw[arr] (s1) -- (crop);
\draw[arr] (crop) -- (s2);
\draw[arr] (s2) -- (ens);
\draw[arr] (ens) -- (out);
\end{tikzpicture}}
\caption{\textbf{Two-stage pipeline.} Stage 1 predicts a coarse global vertex
set from the complete scene. Stage 2 crops the input around the predicted hull,
refines slots initialized from stage 1, and the ensemble converts multiple
stochastic predictions into one consensus wireframe.}
\label{fig:pipeline}
\end{figure*}

\begin{figure*}[t]
\centering
\setlength{\tabcolsep}{1.5pt}
\begin{tabular}{@{}cccc@{}}
\includegraphics[width=0.235\textwidth,trim=0 0 0 70,clip]{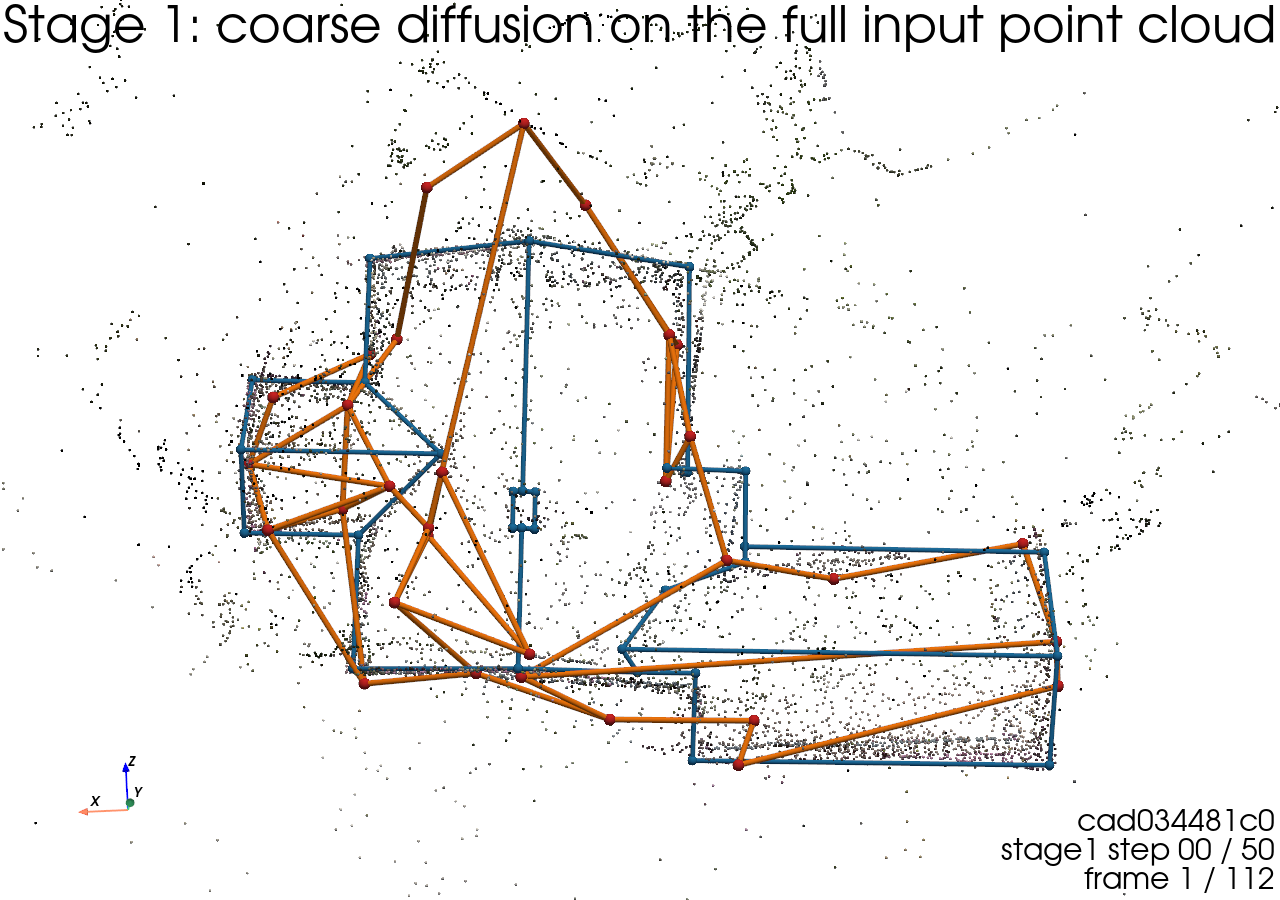} &
\includegraphics[width=0.235\textwidth,trim=0 0 0 70,clip]{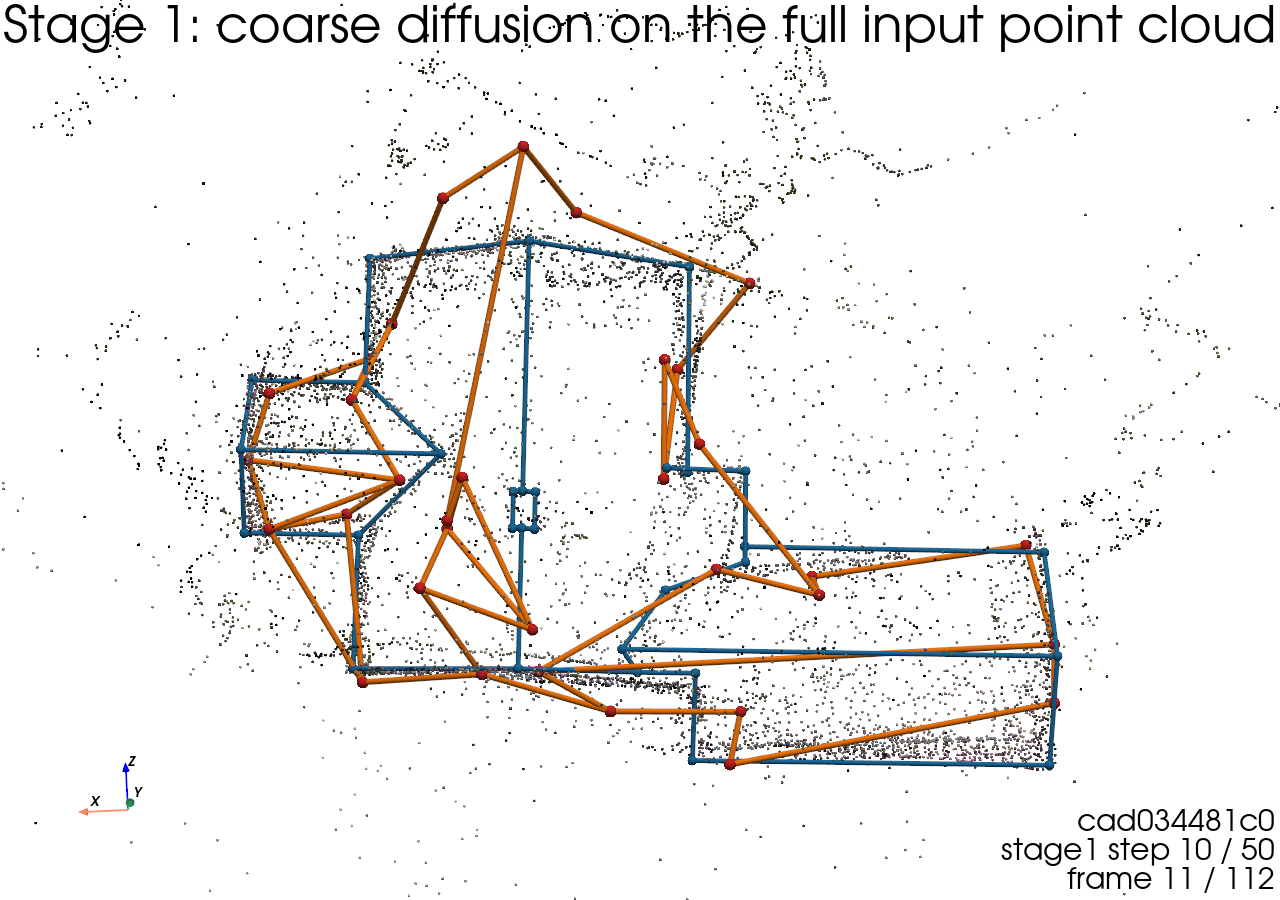} &
\includegraphics[width=0.235\textwidth,trim=0 0 0 70,clip]{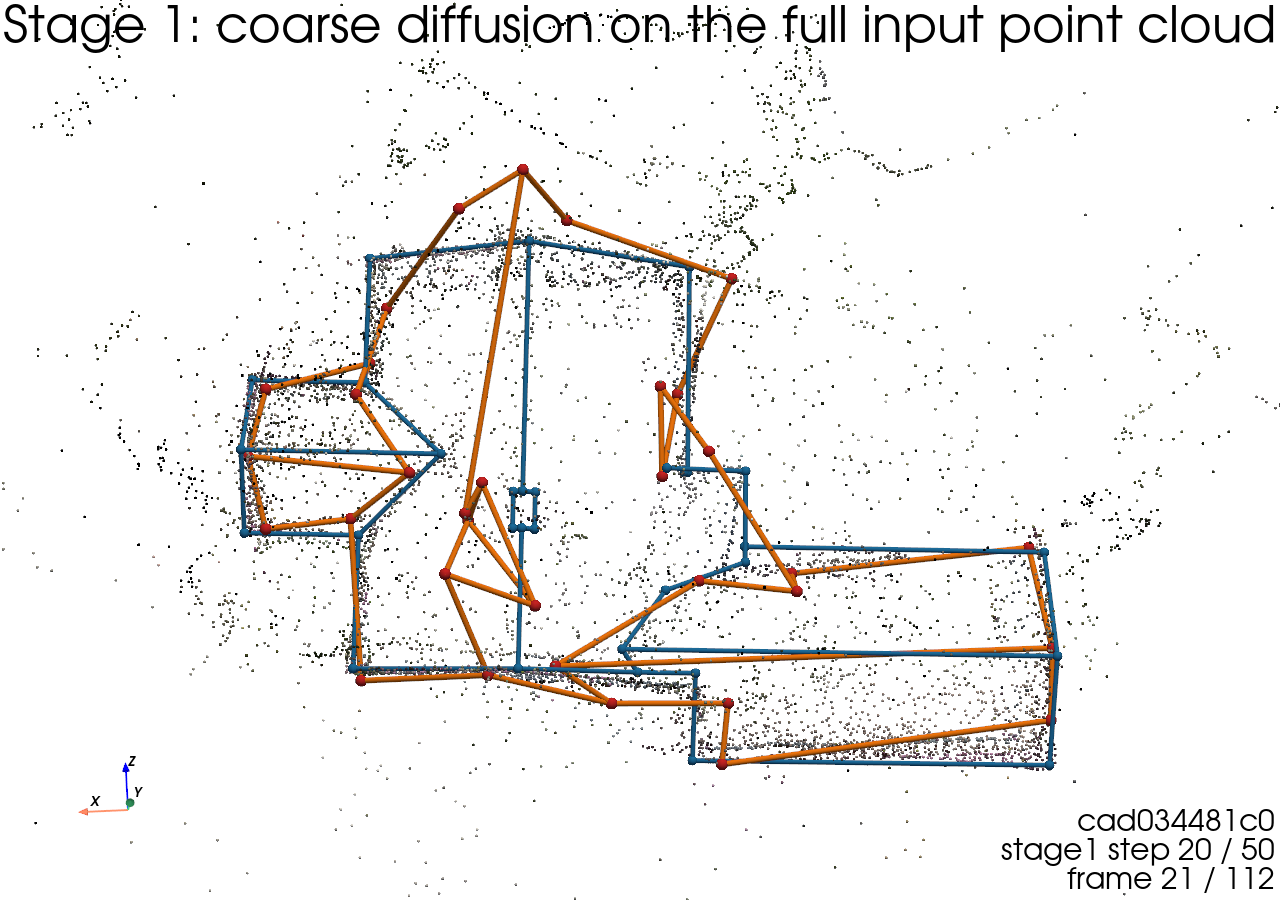} &
\includegraphics[width=0.235\textwidth,trim=0 0 0 70,clip]{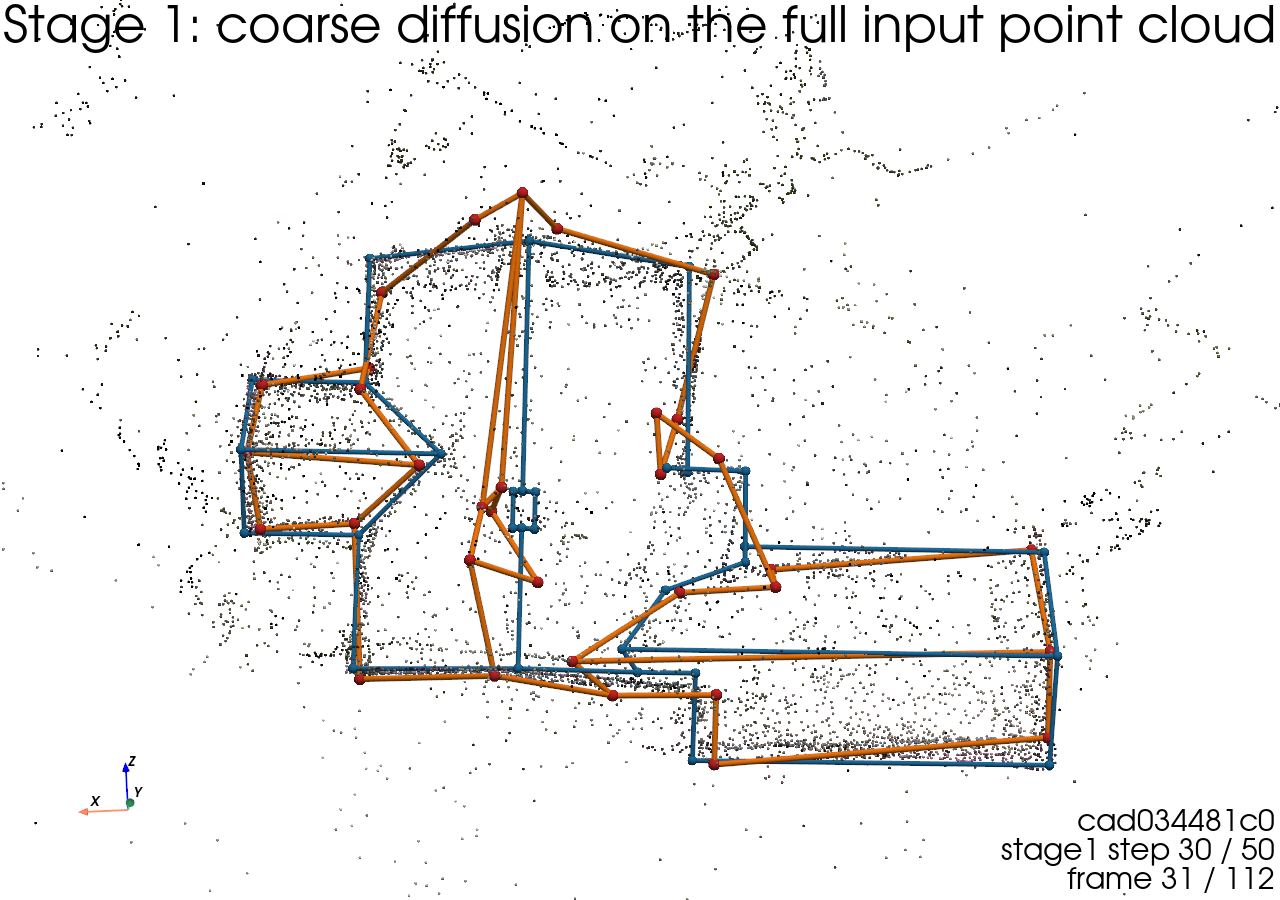} \\
{\scriptsize frame 0000} & {\scriptsize frame 0010} & {\scriptsize frame 0020} & {\scriptsize frame 0030} \\
\includegraphics[width=0.235\textwidth,trim=0 0 0 70,clip]{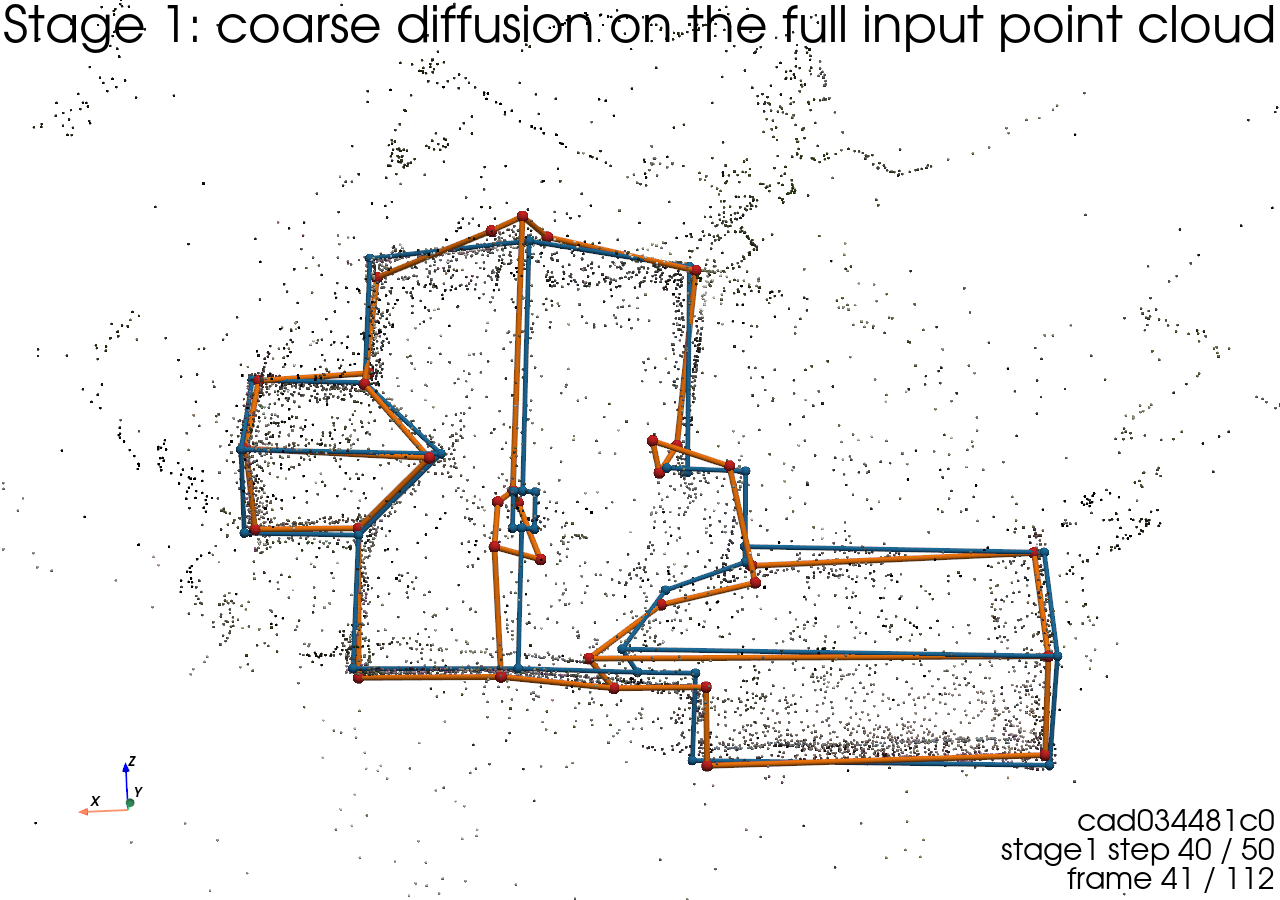} &
\includegraphics[width=0.235\textwidth,trim=0 0 0 70,clip]{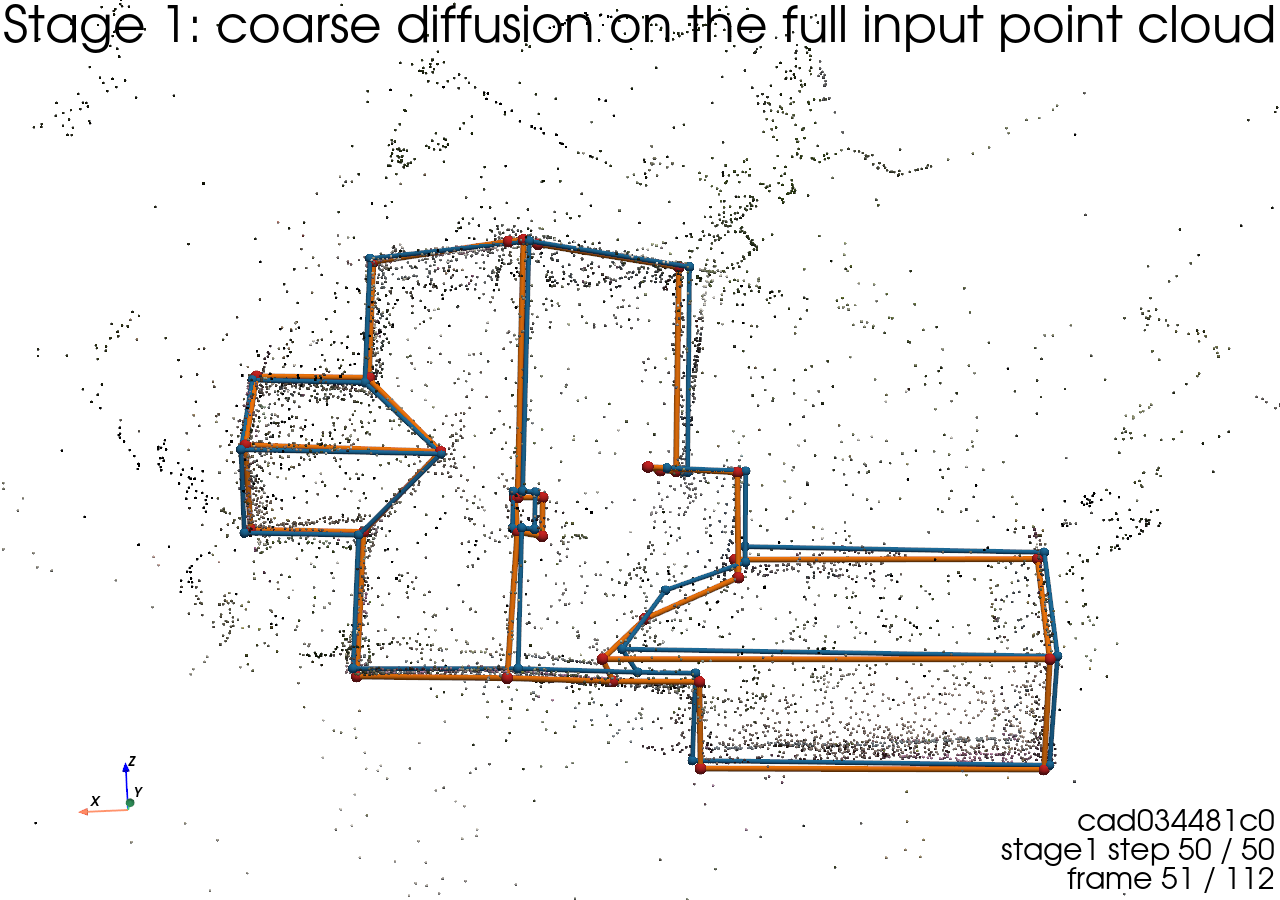} &
\includegraphics[width=0.235\textwidth,trim=0 0 0 70,clip]{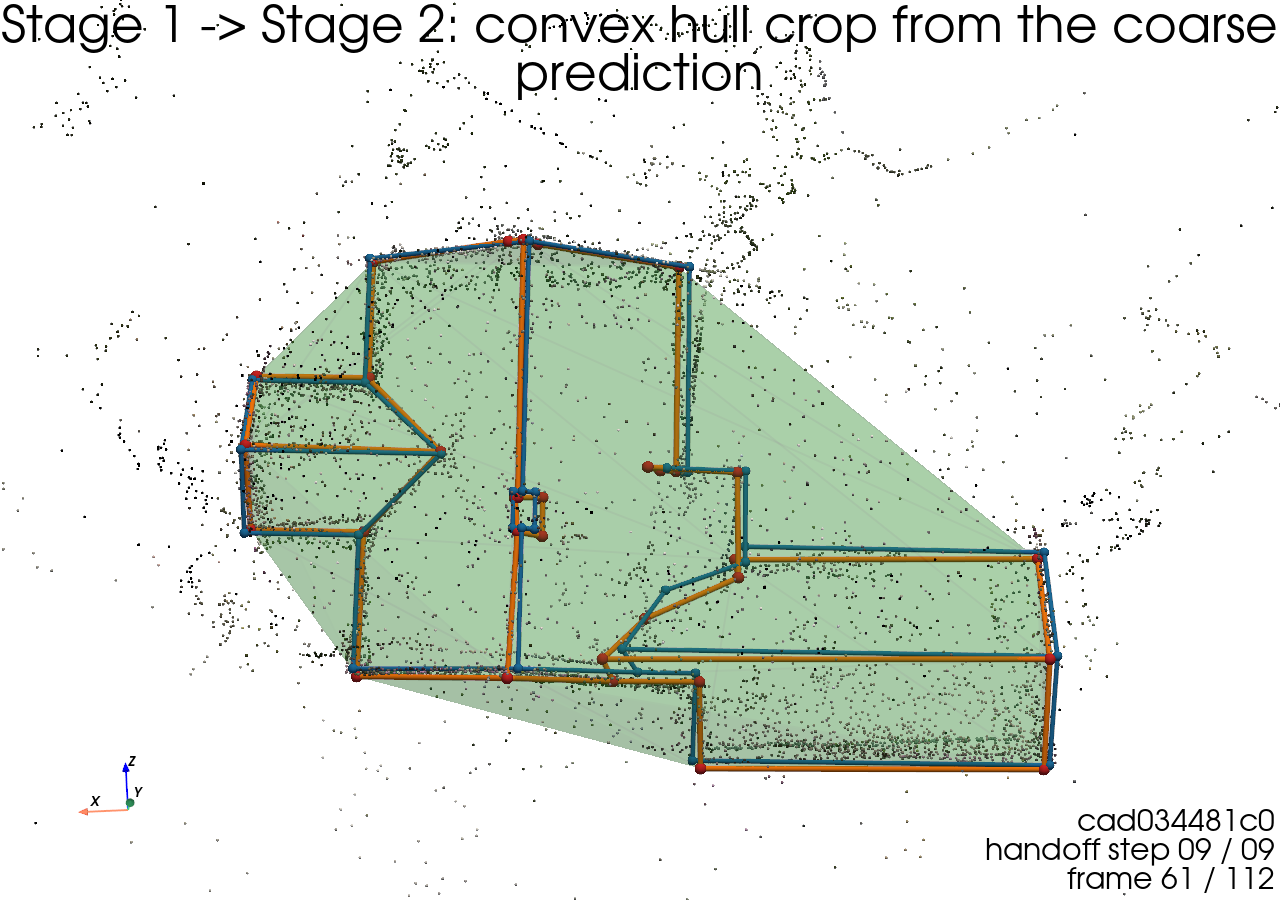} &
\includegraphics[width=0.235\textwidth,trim=0 0 0 70,clip]{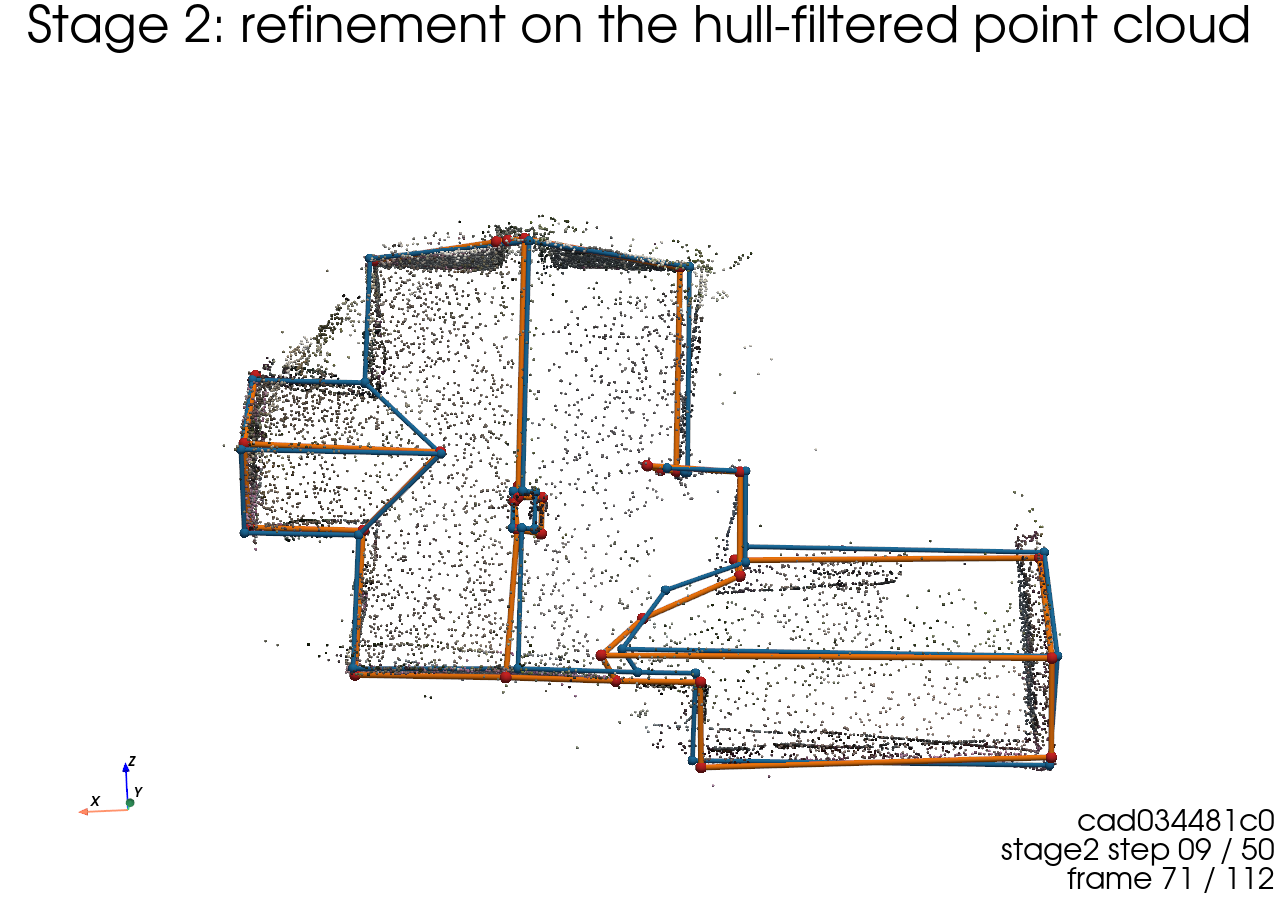} \\
{\scriptsize frame 0040} & {\scriptsize frame 0050} & {\scriptsize frame 0060} & {\scriptsize frame 0070} \\
\includegraphics[width=0.235\textwidth,trim=0 0 0 70,clip]{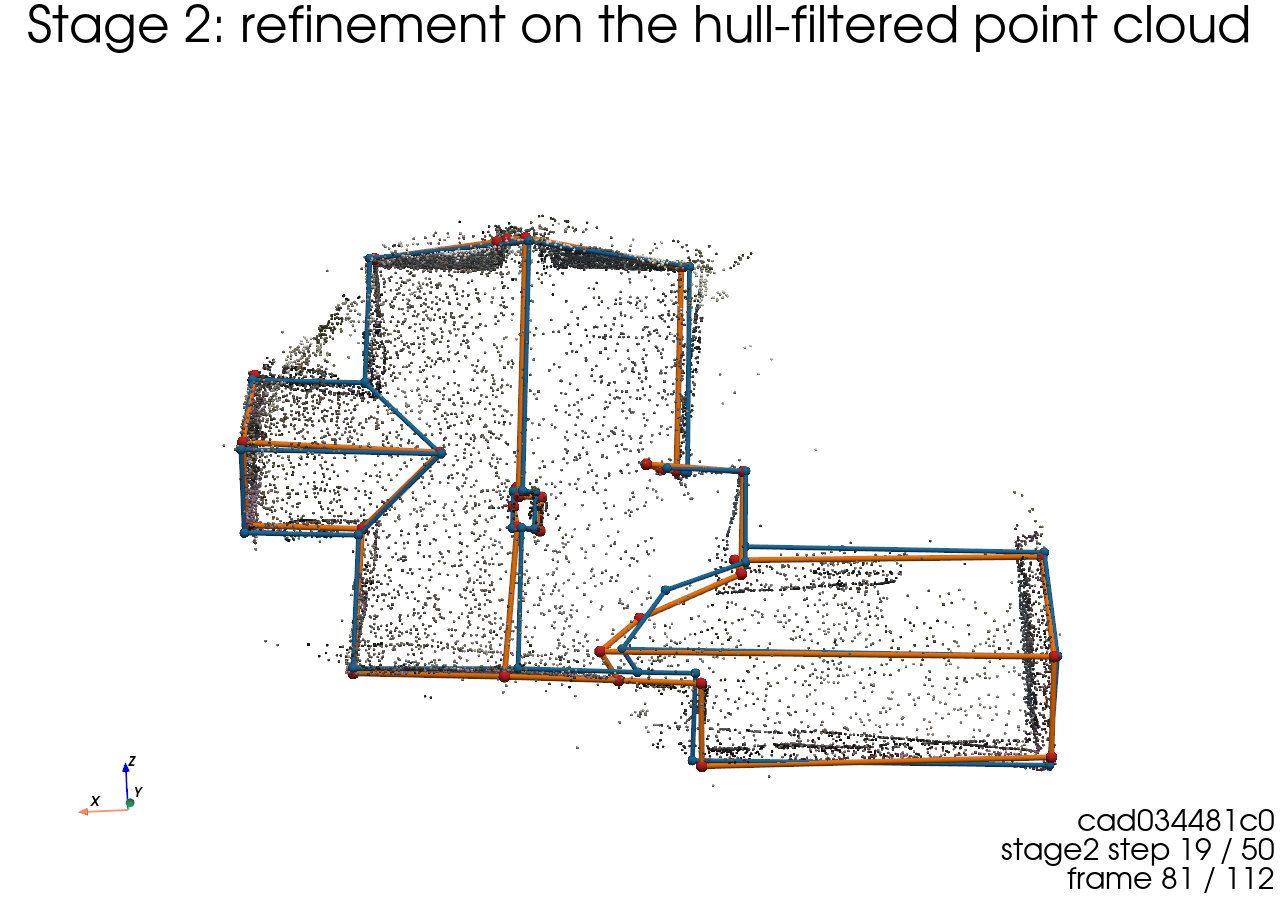} &
\includegraphics[width=0.235\textwidth,trim=0 0 0 70,clip]{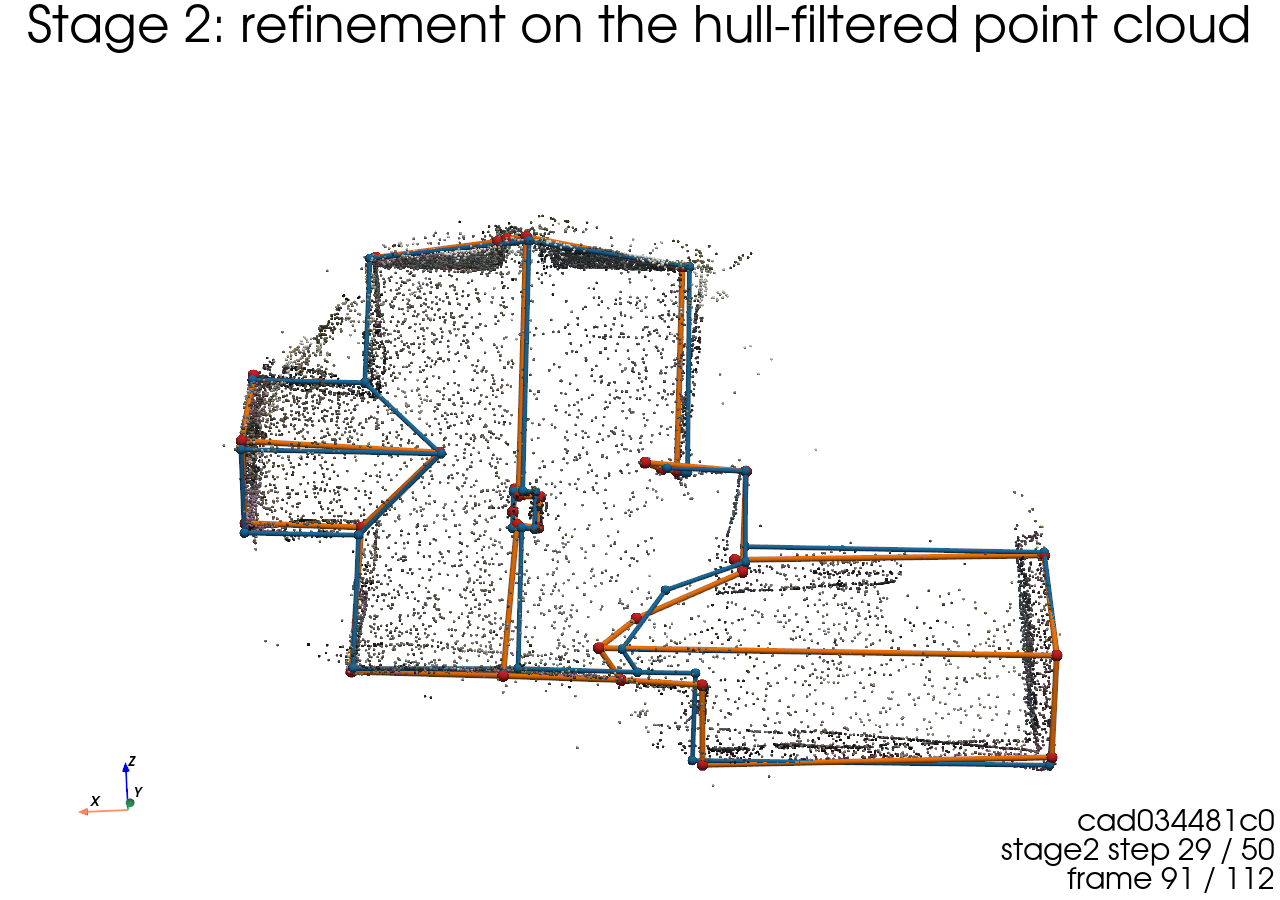} &
\includegraphics[width=0.235\textwidth,trim=0 0 0 70,clip]{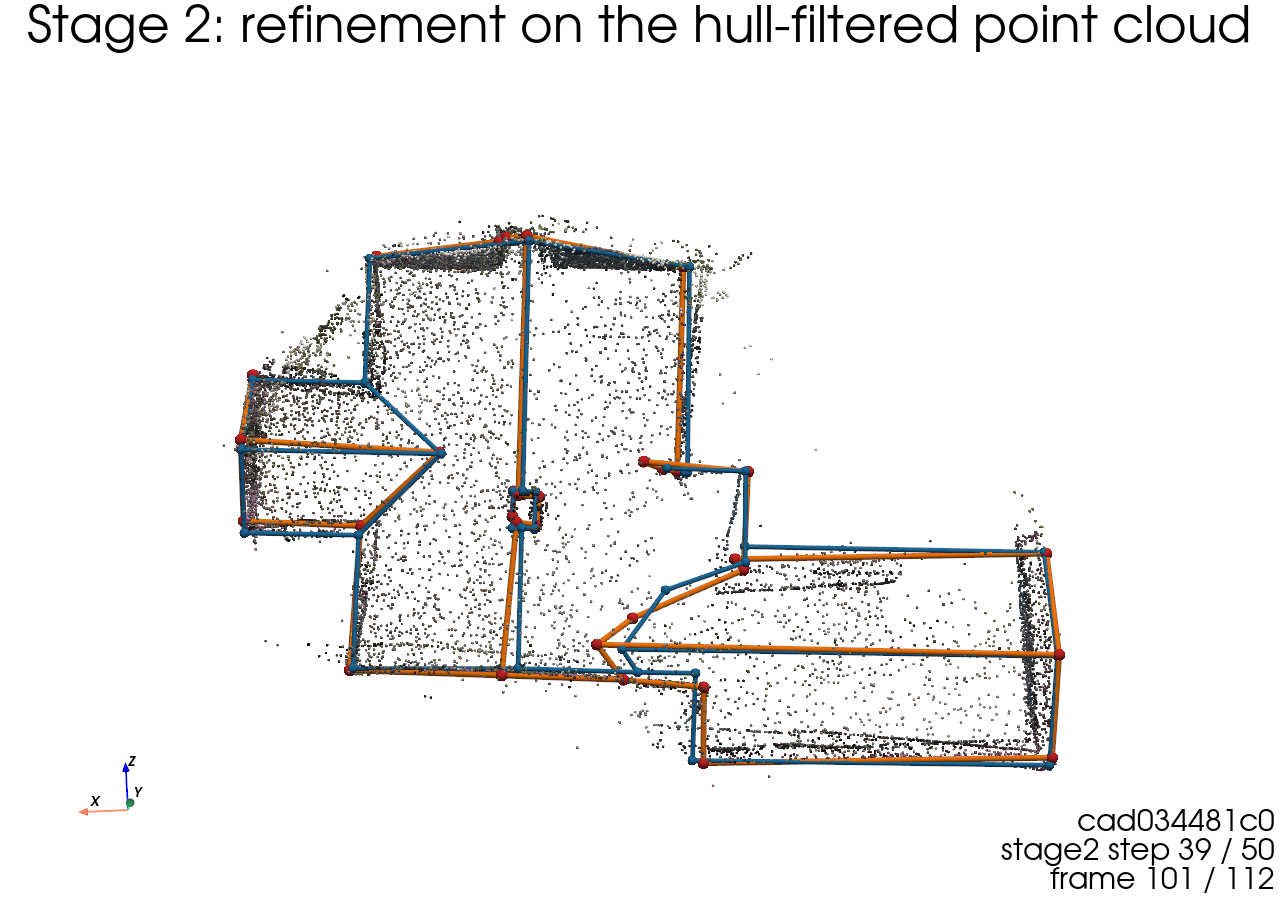} &
\includegraphics[width=0.235\textwidth,trim=0 0 0 70,clip]{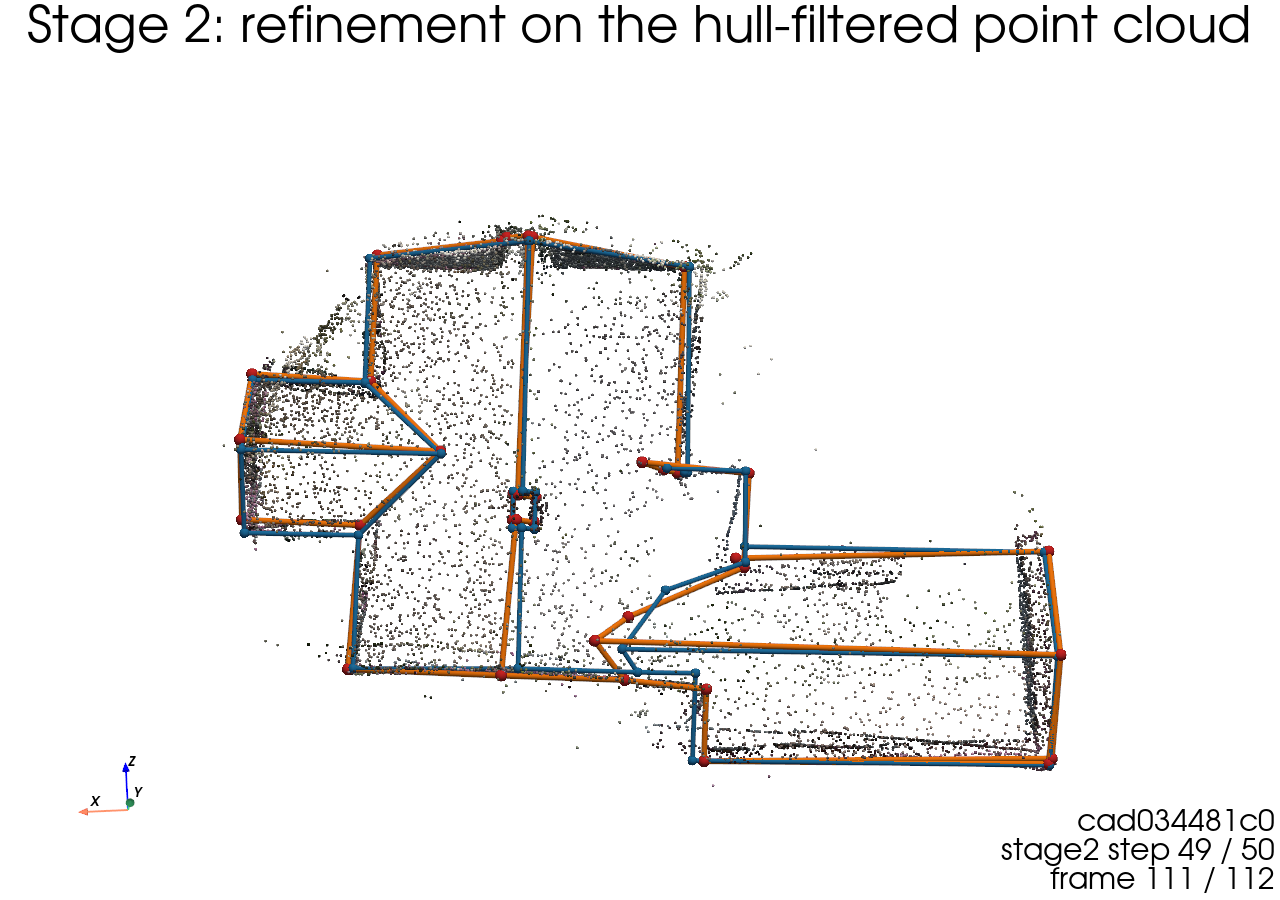} \\
{\scriptsize frame 0080} & {\scriptsize frame 0090} & {\scriptsize frame 0100} & {\scriptsize frame 0110}
\end{tabular}
\caption{\textbf{Denoising trajectory.} Successive visualization frames show
how the pipeline turns scene evidence into a wireframe prediction. Frames
0000--0050 are Stage 1 denoising on the full scene, frame 0060 shows the
convex-hull crop derived from the Stage 1 vertices, and frames 0070--0110 show
Stage 2 refinement inside the cropped scene. The visible point clouds are the
direct inputs to the corresponding scene encoders; the vertex tokens move
through the learned flow under scene-token conditioning before validity and
edge logits produce the final wireframe.}
\label{fig:denoising_frames}
\end{figure*}

\section{Dataset}
\label{sec:dataset}

The challenge is built on the HoHo dataset; the public release
(\texttt{hoho22k\_2026\_trainval}) provides $\approx\!22{,}000$ training and
$\approx\!200$ validation scenes, each a distinct residential building. For
privacy reasons the raw RGB images are not released; each scene is described by an Structure-from-Motion (SfM), Segmentation and Depth Estimation:
a sparse 3D point cloud ($\approx\!40{,}000$ COLMAP
points) with camera poses and intrinsics; per-view metric depth
maps (scale-fit against the COLMAP points) that back-project into
additional 3D points; per-view semantic segmentations from a
roof-specific Gestalt classes (apex, ridge, hip, valley, eave, rake,
fascia, flashing, \ldots) and the general-purpose ADE20k classes; and the
ground-truth 3D wireframe for supervision. Given the nonstandard input, there are not many prior works tangling the same task.

The data is both diverse and noisy. Roof designs range from simple gable and
hip roofs to mansard, multi-gable, and highly articulated structures: wireframes
contain $\approx\!6$--$36$ vertices (mean $22$) over building bounding boxes of
$\approx\!20$--$27\,$m diagonal. Each scene is covered by only $4$--$22$ views
(mean $9$), SfM coverage is sparse on textureless surfaces, and monocular metric
depth drifts in scale and at depth discontinuities. Depth is also frequently
absent (i.e., $\approx\!42\%$ of training and $\approx\!69\%$ of validation
views are pose-only with no depth map). This combination of structural diversity and
input noise, is what makes the task hard and
is the main reason brittle 2D-to-3D lifting baselines struggle.

\section{Method}
\label{sec:method}

We predict vertices as a conditional set and infer edges from the same slots. The
model is applied in two stages: a global generator over the complete scene
followed by a hull-cropped refinement pass initialized from the first prediction
(\cref{fig:pipeline}). Both stages use the same architectural template: a
Perceiver-style scene encoder~\cite{jaegle2021perceiver} produces context tokens (\cref{fig:encoder}), and
a transformer denoiser moves $K=64$ vertex slots through a flow-matching ODE~\cite{lipman2022flow}
(\cref{fig:denoiser}). \Cref{fig:denoising_frames} visualizes this denoising
process over successive inference frames.

\subsection{Stage 1: Global Vertex Generation}
\label{sec:stage1}

\paragraph{Semantic point cloud.}
The network does not use plain point coordinates. Each COLMAP point, depth point,
and camera centre is represented by its normalized $xyz$ position, a Fourier
encoding of that position, a point-type embedding (either COLMAP, depth, or camera centre)
, Gestalt top-1/top-2 semantic
class embeddings, an ADE20k class embedding, and geometric and semantic
confidence scalars. Gestalt and ADE labels for COLMAP points are obtained by
projecting the point into the segmented views and voting over a small image
neighbourhood; depth points inherit the label of their source pixel. Coordinates
are centered by the median COLMAP location and scaled by a robust
95th-percentile radius.

\paragraph{Point subsampling.}
The stage-1 input is a fixed budget of $8192$ points. We sample randomly within
priority tiers. The tiers keep
camera centres and semantically important roof classes first, then fill the
budget with remaining labelled and unlabelled COLMAP/depth points. 
Half of the point budget is reserved for COLMAP and the rest for depth points.

\subsection{Scene Encoder}
\label{sec:scene_encoder}

The encoder, shown in \cref{fig:encoder}, first maps each enriched point to a
$d$-dimensional feature. Two full self-attention layers contextualize all input
points. A fixed set of 1024 learned pooling tokens is then anchored at scene
points selected from priority Gestalt classes and cross-attends to the full point
set; two self-attention layers refine the scene tokens. In the submitted system
both stages use the large configuration, $d=320$ with 10 attention heads.

\subsection{Vertex Denoiser and Losses}
\label{sec:denoiser}

The denoiser is a DiT-style transformer~\cite{peebles2023dit} over $64$ vertex tokens. At flow time
$t\in[0,1]$, each token contains a current 3D position. The token position and its Fourier
features are projected to the model width, a sinusoidal time embedding is used
for adaptive LayerNorm modulation, and each DiT block cross-attends to the time
token and scene tokens. The heads predict a velocity in $\mathbb{R}^3$, a validity logit, and pairwise edge
logits. Stage 1 and stage 2 both use 12 denoiser blocks.

Training uses flow matching~\cite{lipman2022flow} with the linear interpolant
\begin{equation}
  x_t = (1-t)x_0 + tx_1, \qquad v^\star = x_1-x_0 ,
\end{equation}
where $x_0$ is a noisy slot initialization and $x_1$ is a ground-truth vertex.
Ground-truth vertices are assigned to slots by Hungarian matching, giving
permutation-invariant supervision. The objective is a weighted sum of SmoothL1
velocity loss, SmoothL1 endpoint loss, focal validity loss for real/null slots,
and pairwise BCE edge loss:
\begin{equation}
  \mathcal{L} =
  \mathcal{L}_{flow}
  +0.5\mathcal{L}_{end}
  +0.2\mathcal{L}_{valid}
  +0.2\mathcal{L}_{edge}.
\end{equation}
Null slots are retained with a smaller weight so the model learns both where
vertices are and when a slot should be inactive. At inference the learned ODE is
integrated with 50 Euler steps; a representative trajectory is shown in
\cref{fig:denoising_frames}.

\subsection{Stage 2: Hull-Cropped Refinement}
\label{sec:stage2}

Stage 2 receives the stage-1 vertices as its initialization, as summarized in
\cref{fig:pipeline}. We build the convex hull of the predicted vertices, inflate
it by a small metric margin, and resample
a denser point cloud inside this hull. The crop removes far-field clutter while
preserving roof evidence near the predicted structure. Points are now sampled
randomly from COLMAP and depth-derived points, using a larger budget of $16384$
points. The cropped scene is encoded by the same encoder design, and the
denoiser refines the stage-1 vertex slots for another 50 Euler steps. Thus stage
1 solves the global localization problem, while stage 2 spends more capacity on
metric refinement around a strong initial wireframe.

\paragraph{Ensemble inference.}
The submitted system runs 16 stochastic trajectories, with randomness coming
from slot initialization. The stage-2 crop is built from the union of stage-1
predictions, allowing the 16 refinements to share one encoded crop. We select
the medoid prediction under a Hungarian-aligned vertex distance plus edge
agreement, then replace each selected vertex by a consensus aggregate of its
matched vertices across the other samples. This strongly reduces run-to-run
variance without changing the overall topology.

\section{Experiments}
\label{sec:experiments}

\paragraph{Training details.}
We trained on the official training split using cached semantic point clouds and
evaluated on the public validation split. We also removed 115 training scenes
whose preprocessing sanity score was below $HSS=0.05$, eliminating a small
set of corrupted or poorly aligned frames before model training. Stage 1 used
the large model throughout ($d=320$, 12 denoiser blocks, 34M parameters) and
was trained for 2000 epochs on 4 H200 GPUs with DDP. The per-GPU batch size was 32
(global batch 128), with bf16 autocast, bf16-compressed gradient all-reduce,
fused AdamW (weight decay $10^{-4}$), gradient clipping at 1.0, and a
1000-step linear warmup followed by cosine decay. The peak learning rate in the
used large stage-1 run was $5\cdot10^{-4}$.

Stage 2 used the same large model size and was trained on hull-cropped cached
scenes initialized from stage-1 predictions. The selected random-sampling
checkpoint was fine-tuned for 1000 epochs at $10^{-5}$ from the best stage-2
checkpoint, again on 4 H200 GPUs with global batch 128, bf16, DDP, gradient
clipping, and the same loss weights as in \cref{sec:denoiser}. Inline validation
during training used 20 stage-2 sampling steps for speed; final inference used
50 steps in both stages.

\begin{table}[t]
\centering
\small
\resizebox{\columnwidth}{!}{%
\begin{tabular}{lccccc}
\toprule
Configuration & Val. HSS & Preproc. & Stage 1 & Stage 2 & Total \\
\midrule
Stage 1 large & 0.4764 & 2.56s & 0.58s & -- & 3.14s \\
Stage 1 + Stage 2 & 0.4977 & 2.56s & 0.57s & 1.17s & 4.30s \\
16-sample ensemble & 0.5025 & 2.56s & 0.58s & 2.38s & 5.52s \\
\bottomrule
\end{tabular}
}
\caption{\textbf{Validation sanity results.} Scores are local validation
HSS from the submission sanity checks or checkpoint metadata. Preprocessing is
the shared point-cloud construction time, Stage 1 reports encoder+denoiser time,
and Stage 2 reports hull-crop refinement time. The ensemble timing uses batched
16-sample inference with 50/50 sampling steps.}
\label{tab:experiments}
\end{table}

The final ensemble adds a modest but consistent gain over a single two-stage
sample, and more importantly makes inference substantially less sensitive to the
random slot initialization. Medoid selection removes outlier trajectories, while
cross-sample consensus stabilizes vertex positions across runs.

\section{Conclusion}
\label{sec:conclusion}

We presented a two-stage conditional set generator for S23DR wireframe
reconstruction. The method enriches sparse SfM and depth points with semantic
features, encodes the scene with a Perceiver-style transformer, and denoises a
fixed set of vertex slots with a flow-matching DiT. A global first pass predicts
coarse vertices, a hull-cropped second pass refines them, and a lightweight
multi-sample consensus ensemble strongly reduces run-to-run variance. This
architecture was the winning S23DR 2026 submission, ranking first on the private
leaderboard.


\bibliographystyle{ieeetr}
\bibliography{refs}

\end{document}